\documentclass{article} %
\usepackage{iclr2027_conference,times}
\iclrfinalcopy

\usepackage{amsmath,amsfonts,bm}

\def\eqref#1{equation~\ref{#1}}

\def\1{\bm{1}}

\DeclareMathAlphabet{\mathsfit}{\encodingdefault}{\sfdefault}{m}{sl}
\SetMathAlphabet{\mathsfit}{bold}{\encodingdefault}{\sfdefault}{bx}{n}

\usepackage{hyperref}
\usepackage{url}

\usepackage{xspace}
\usepackage{graphicx}
\usepackage{listings}
\usepackage{wrapfig}
\usepackage{booktabs}
\usepackage{tabularx}

\title{Lost in Translation: Measuring the Effect of Non-Native English on End User Performance of Large Language Models}

\author{Yusheng Zhou\thanks{Equal contribution.}\hspace{1.5em}Eleanor Lin\footnotemark[1]\hspace{1.5em}David Jurgens \\
Electrical Engineering and Computer Science Department\\
University of Michigan\\
Ann Arbor, MI 48105, USA \\
\texttt{\{yszhou,elealin,jurgens\}@umich.edu}
}

\newcommand{\myparagraph}[1]{\paragraph{#1}}

\newcommand{\tref}[1]{Table~\ref{#1}}

\newcommand{\dataset}{{\textsc{Fable}}\xspace}
\newcommand{\numofdata}{{190,911}\xspace}

\begin{document}

\maketitle

\begin{abstract}
Large language models (LLMs) are increasingly used by people whose first language is not English, yet these users have been shown to receive systematically lower-quality responses than fluent speakers. Which specific features of non-native English drive this gap remains unclear, because fluency is itself a composite of mechanical accuracy, vocabulary use, organization, and discourse coherence. Here, we introduce \dataset, a controlled dataset of \numofdata English prompt variants derived from 174K real user prompts for writing-related tasks. Evaluating responses from 34 open-weight LLMs, we find a clear asymmetry; while models do not propagate surface errors such as misspellings into their outputs, models do mirror higher-level rhetorical and lexical qualities present in the user's prompt. Further, the overall quality of responses differs substantially between the least- and most-fluent prompts. These results highlight a key LLM performance disparity for non-native English LLM users, resulting in both lower-quality and less-fluent answers.
\end{abstract}

\section{Introduction}
Non-native English speakers represent a critical and rapidly growing user base for large language models (LLMs) \citep{liu2025ai}. However, LLMs have demonstrated performance degradation for non-native English speakers \citep{reusens2025native}. Intuitively, this performance degradation may stem from differences in the language produced by non-native speakers of different fluency levels, compared to the native English that LLMs are trained on. Fluency, however, is not a single property: it is a composite of features ranging from surface mechanics such as spelling and grammar to higher-level qualities such as organization and discourse coherence \citep{CHAMBERS1997535}. Without knowing which of these features matter for LLM behavior, it is difficult to diagnose why non-native speakers are disadvantaged and to design effective interventions to improve LLM performance for these users. Here, we systematically test which features of non-native English drive performance degradation for non-native speakers. 

Existing work on prompt sensitivity has examined how surface-level variations such as formatting \citep{he2024does}, example ordering \citep{guo-etal-2024-makes, lu2022fantastically}, length \citep{liu2025effectspromptlengthdomainspecific}, tone \citep{dobariya2025mind}, and politeness \citep{yin2024should} affect model behavior. A smaller line of work has begun to consider specific linguistic properties such as tense, mood, and voice \citep{leidinger2023language} and the broad effect of speaker nativeness \citep{reusens2025native}. However, these studies typically treat fluency as a single binary or scalar property and rely on benchmark-style prompts that do not reflect how everyday users actually interact with LLMs. As a result, it remains unclear which specific aspects of non-native English are most consequential for downstream task performance, or whether their effects are consistent across realistic, open-ended user tasks.

We address this gap by introducing \dataset (Fluency-Adjusted Benchmark for LLM Generation Evaluation), a controlled dataset of \numofdata English prompt variants derived from real-world user interactions for writing-related tasks. Starting from 174K filtered prompts in English and Chinese collected from WildChat \citep{WildChat} and ShareLM \citep{ShareLM}---a language pair representing one of the largest populations of bilingual LLM users \citep{yang2006learners}---we generate multiple English translations of each Chinese prompt that conform to specified profiles along the five dimensions of the ESL Composition Profile \citep{ESL}: Content, Organization, Vocabulary, Language Use, and Mechanics. This design disentangles the contribution of each fluency dimension while holding the prompt's semantic intent approximately constant. We evaluate 34 open-weight LLMs on a shared sample of 5,000 prompt variants, and evaluate their responses using both rule-based linguistic metrics and LLM-as-judge ratings of fluency and overall response quality.

Our analysis reveals a striking asymmetry between Surface Form Features and higher-order qualities. LLMs do not substantially propagate surface errors (e.g., misspellings, grammatical mistakes) from prompts into their responses, but they do mirror higher-level rhetorical and lexical qualities: prompts with weaker Content, Organization, or Vocabulary yield responses with correspondingly weaker proficiency along those same dimensions. More importantly, prompt fluency has a substantial effect on response quality: a one-point increase in each retained prompt-fluency dimension on the four-point scale leads to a 0.04--0.14 point increase in overall response quality on a five-point scale. This effect is consistent across writing-focused task categories, although the relative importance of individual fluency dimensions varies by task. Together, these findings suggest that users who turn to LLMs to compensate for limited English fluency may still receive systematically lower-quality output, with implications for equitable access to LLM technology.

Our paper makes the following three contributions. (1) We introduce \dataset, a controlled dataset of \numofdata prompt translations spanning the ESL Composition Profile that enables fine-grained study of how prompt fluency affects LLM behavior. (2) We provide an evaluation methodology that jointly quantifies surface-level and higher-order linguistic properties of both prompts and responses. (3) We demonstrate empirical evidence that LLM response quality is shaped more by the rhetorical and lexical qualities of a prompt and that, across 34 models, LLM responses to less-fluent prompts are themselves less fluent and lower quality. Our results point to an important source of inequality in access to fluent, high-quality outputs by all users. All data and code will be made available upon publication (CC-by-SA-4.0).

\section{Related Work}

\myparagraph{Language Fluency and Language Technology} 
Language fluency has been shown to play a critical role across multiple NLP fields. In Machine Translation, prior work demonstrates that noisy input significantly degrades translation quality for both neural MT systems and large language models \citep{popovic2024effects, pan2024can}. This phenomenon can be viewed as a manifestation of reduced language fluency of input. Meanwhile, ``fluency" is also recognized as a fundamental dimension in the evaluation of translation quality, with researchers proposing structured frameworks, such as MQM \cite{park2024multi}, to systematically assess fluency alongside other aspects of translation quality. Similarly, in Information Retrieval, the effectiveness of retrieval systems is closely tied to the quality of user queries: well-formed queries tend to achieve better performance \cite{chikkamath2024your}, while typos or other linguistic errors can substantially reduce system accuracy \cite{zhuang2022characterbert}. In this work, we seek to understand the impact of language fluency on LLM responses to non-native English speakers.

\myparagraph{Native v. Non-native Use of LLMs}
Prior studies about contrastive rhetoric have shown that first language and cultural background influence how people write in a second language \cite{kaplan1966cultural}. In particular, native and non-native English speakers show systematically different usage of linguistic structure beyond semantic content \cite{rabinovich2016similarities}. Such variation naturally carries over to interactions with LLMs; e.g., \citet{reusens2025native} demonstrate that the nativeness of prompts significantly influences model performance. Our work identifies the specific aspects of non-native prompt fluency that cause differences in model behavior.

\myparagraph{Effects of Prompt Design}
Prompts that are semantically equivalent can exhibit a substantial model performance gap solely due to variations in their surface form \citep{cao2024worst}. Even when the semantic content remains identical, different formatting can lead to different performance \citep{he2024does, sclar2023quantifying}. Example ordering also plays an important role in designing prompts for in-context learning \citep{lu2022fantastically, guo-etal-2024-makes}. Prompt length also serves as a decisive factor in domain-specific tasks \citep{liu2025effectspromptlengthdomainspecific}. Additionally, in real-world interactions, research indicates that the tone \citep{dobariya2025mind} and politeness \citep{yin2024should} of a prompt significantly influence how models respond. This extensive evidence of LLM sensitivity to variation in prompts' surface form provides the motivation for our own comparison of semantically equivalent prompts at different fluency levels. %

While these studies provide valuable insights into different dimensions of prompts, a systematic evaluation of how the language fluency level of prompts influences model behaviors is still missing. Moreover, prior studies often evaluate prompt effectiveness using benchmark tasks that may not reflect real-world user interactions. In contrast, our work systematically investigates how variations in prompt fluency influence LLM responses on real-world interaction data.

\section{Curating a Dataset of Realistic, Translated User Prompts}
\label{sec:data}

To investigate the impact of fluency in prompts on model responses and downstream task performance, we curate a set of prompts from real-world interactions in the WildChat ~\citep{WildChat} and ShareLM ~\citep{ShareLM} datasets. Both datasets consist of conversations of diverse domains collected from real human-AI interactions. These prompts allow us to better match the experience of everyday users in realistic workflows (e.g., summarizing an email) when compared with prompts for multiple choice questions traditionally seen in NLP.

\begin{figure}[t]
  \includegraphics[width=0.95\columnwidth]{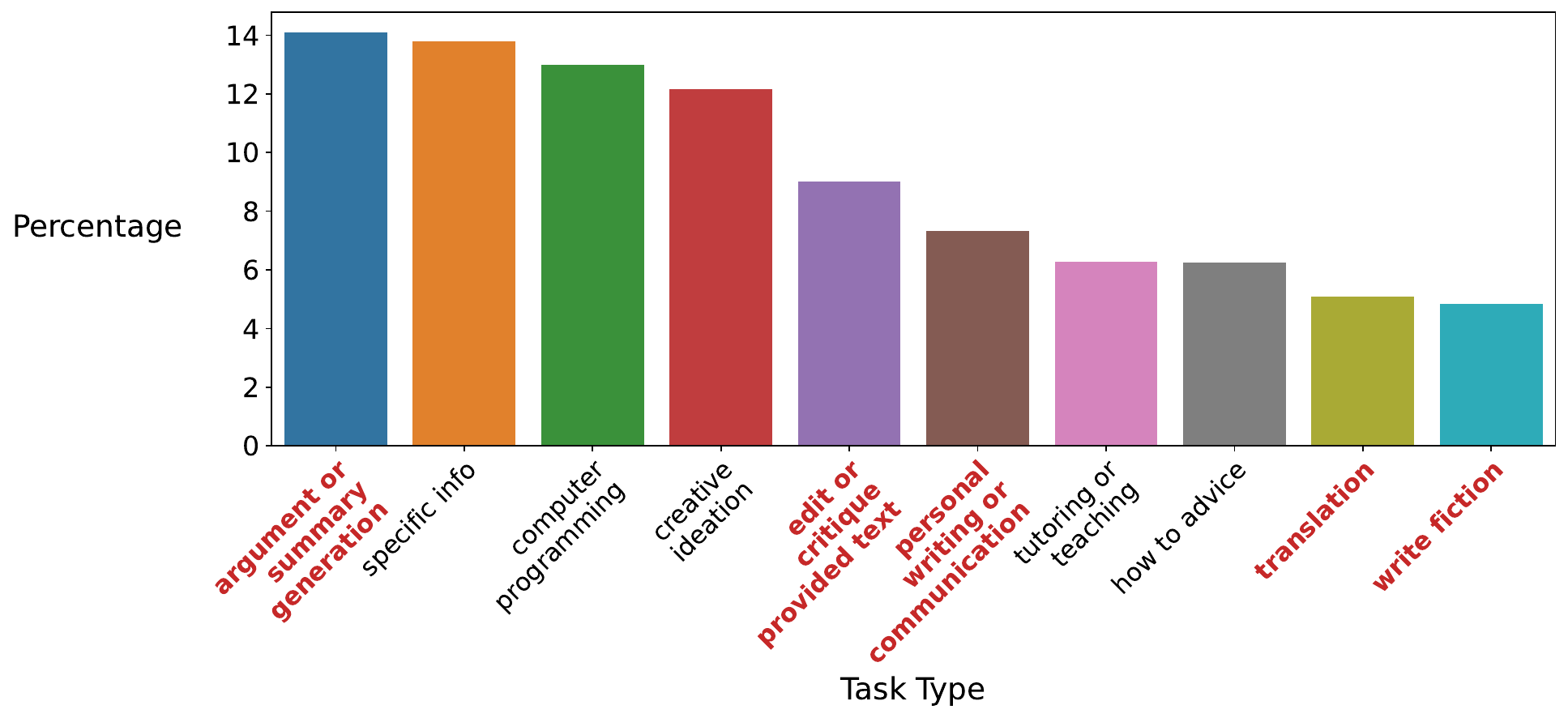}
  \caption{The top-10 task categories for Chinese prompts in WildChat and ShareLM,  categorized using the taxonomy proposed by \citet{chatterji2025howchatgpt}, shows that many real-world user prompts involve writing and information-seeking tasks rather than other questions. Bold denotes categories of prompts we use. }
  \label{fig:task_type_distribution}
\end{figure}

\myparagraph{How Are People Using LLMs?}
Adapting the methodology and taxonomy proposed in ~\citet{chatterji2025howchatgpt}, we categorize each prompt into one of 24 predefined task categories. The distribution of prompts across task categories is shown in Fig~\ref{fig:task_type_distribution}. We observed that among the ten most common user task categories, writing-related queries account for the majority, with ``Computer Programming" being the only category not directly related to writing. Moreover, we believe that LLM responses in these writing-related tasks such as “argument or summary generation,” “creative ideation,” and “edit or critique provided text” may be influenced by users' fluency.

\myparagraph{Data Filtering} To ensure data quality and experimental consistency, we apply a sequence of data filtering steps. For multi-turn conversations, we only keep the first user prompt. While we recognize the importance of multi-turn conversations, our focus here is on how differences in linguistic proficiency affect the initial model response, rather than whether users can \textit{eventually} achieve comparable outcomes to native speakers after multiple rounds of repair. The first-turn model response is consequential because it determines what the user must subsequently correct, clarify, or refine. A systematic quality deficit at the first turn means that non-native users begin the interaction at a disadvantage and must expend additional turns, time, and effort to reach parity; this additional interaction burden is itself an equity cost. Additionally, effectively simulating realistic interactions across multiple turns remains a challenging problem \citep{ivey2024real}, and there are currently no curated datasets that are easily amenable to testing multi-turn effects of fluency. Therefore, we leave exploration of multi-turn scenarios to future work.

Additionally, since it is difficult to evaluate the fluency of very short prompts, and extremely long prompts are less representative of typical user queries, we perform length-based filtering. We remove both the shortest and longest 5\% of prompts and randomly retain 30\% of the prompts shorter than 30 tokens to control the prompt length distribution. 

To filter out prompts that contain multiple languages, we perform sentence-level language identification using the fastText language identification model ~\cite{fasttext}. We segment each prompt into sentences and keep prompts which have all sentences classified as the same language with confidence scores above 0.6. 

In the following experiments, we focus on prompts in English and Chinese. English is the dominant language used in LLM training and inference \citep{qin2024multilinguallargelanguagemodel}, while Chinese represents one of the most-spoken non-English languages used with LLMs \citep{liu-etal-2024-openeval}. Further, Chinese-English bilinguals worldwide make up a large, natural population of speakers with varying degrees of English fluency who are therefore affected by models' sensitivity to English fluency. Studying these two languages also reflects common real-world scenarios where bilingual users may tend to interact with LLMs in English even when it is not their first language \citep{10.1145/3772363.3798492}. We do recognize that non-native English speakers have many other first languages, each producing characteristic interference patterns \citep{kobayashi1992effects}. The specific fluency effects we observe may not generalize to other source languages, particularly those with substantially different typological properties (e.g., morphologically rich languages, or languages with writing systems beyond Han characters). Nonetheless, the fluency taxonomy we use is language-agnostic. We therefore expect aspects of our findings to generalize beyond Chinese-speaking users.

We restrict the prompts to those focusing on writing, as these represent the most common forms of real-world LLM usage according to our analysis of WildChat and ShareLM and prior studies on real-world usage data, e.g. \citet{chatterji2025howchatgpt}. Specifically, we only include prompts from the following task types: ``argument or summary generation," ``edit or critique provided text," ``personal writing or communication," ``translation," and ``write fiction." These prompts are well-suited for studying the relationship between prompt fluency and model behavior. 
After the above processing steps, our final dataset consists of 173,726 prompts. WildChat contributes 21,734 English and 27,532 Chinese prompts, while ShareLM contributes 90,121 English and 34,339 Chinese prompts.
To provide a reference for downstream response quality evaluation, we use DeepSeek-V4.1-Flash \citep{deepseekai2026deepseekv41flash} to translate the original Chinese prompts into English, and treat the resulting outputs as ground-truth translations.

\section{Introducing Translation Artifacts}

Language fluency is a multidimensional spectrum, with many features of language contributing to the perception of fluency, such as grammatical structure, word choice, discourse coherence, or even capitalization. To more precisely control for these different factors, we introduce \dataset, which systematically constructs prompts using heterogeneous sources of disfluencies and provides multiple variants of the same prompt intent across the fluency spectrum. \dataset allows testing how specific types of linguistic behavior influence a model's downstream behavior. Here, we first describe how we measure fluency and then how we generate the data for \dataset.

\subsection{Quantifying Fluency}
\label{Fluency Measurement}

We quantify language fluency from two complementary perspectives: \textbf{Surface Form Features} that measure specific properties like mechanical competence (spelling, capitalization), and \textbf{Rhetorical and Lexical Qualities} features that measure higher-level constructs like discourse coherence and vocabulary use. Surface Form Features mark the  writer's ability to control the form of the language \citep{witte1981coherence}, while rhetorical and lexical qualities measure compositional competence \citep{siekmann2022structure}.

For \textbf{Surface Form Features}, we build a rule-based evaluation pipeline based on LanguageTool \citep{milkowski2010developing,languagetool} to detect spelling, grammatical, lexical, and stylistic errors. We also compute Flesch Reading Ease to evaluate readability \citep{flesch1948new}. 

For \textbf{Rhetorical and Lexical Qualities}, we adopt a widely used analytic ESL writing rubric which evaluates a given text along five dimensions: Content, Organization, Vocabulary, Language Use, and Mechanics~\citep{ESL}. A brief description of all dimensions is presented in Table~\ref{tab:esl_intro}. We collect these proficiency dimensions using an LLM-as-a-judge framework, where the judging model assigns an integer score from 1 to 4 for each dimension independently according to the rubric definitions. We use \texttt{google/gemma-4-26B-A4B} as judge. To validate our pipeline, we use the ICNALE dataset~\citep{ishikawa2018icnale}, consisting of 656 learner-written texts annotated with four-level CEFR-aligned writer proficiency labels. We apply our judge template and take the average score from all dimensions; this score has a correlation of 0.47, suggesting the judge is moderately well calibrated on what is a hard task for humans.

\begin{table}[]
\centering
\small
\setlength{\tabcolsep}{4pt}
\begin{tabular}{ll}
\toprule
\textbf{Dimension} & \textbf{Description} \\
\midrule
Content      & Topic relevance and depth of idea development \\
Organization & Logical sequencing, coherence, and paragraph structure \\
Vocabulary   & Lexical richness, word choice, and idiomatic usage \\
Language Use & Grammar and sentence construction \\
Mechanics    & Spelling, punctuation, capitalization, and formatting \\
\bottomrule
\end{tabular}
\caption{The five dimensions in the adapted ESL Composition Profile \citep{ESL}.}
\label{tab:esl_intro}
\end{table}

\subsection{Controlled Generation of Translation}

To systematically study the effect of prompt fluency on models' responses, we construct \dataset to contain a range of translations with controlled fluency levels. 
Specifically, we create prompts that instruct according to the five dimensions of \textbf{Rhetorical and Lexical Qualities}. Each dimension is assigned a discrete level from 1 to 4, corresponding to increasing levels of proficiency in this specific dimension: Level 1 denotes \textit{very poor} proficiency, level 2 represents \textit{fair to poor} proficiency, level 3 means \textit{good to average} proficiency, and level 4 stands for \textit{excellent to very good proficiency}. Each combination of these levels defines a \textit{translation profile}. For each reference Chinese-language prompt, we include (i) 10 translations produced by random samples of valid profiles and (ii) translations from four uniform proficiency profiles where all dimensions have the same level. We then use Qwen3-32B to generate translations that conform to the specified ESL composition profile.
Appendix \tref{tab:translation_example} presents an example of translated prompts under different instructed levels. This process yields multiple translation variants per prompt, enabling a systematic exploration of translation-induced prompt variation across different dimensions of fluency. 
 
Because LLM-generated translations may contain extraneous text or
generation artifacts, we apply post-generation quality control before
constructing the final dataset. This procedure combines rule-based
checks for generation-related text, repeated trigram detection,
language identification, and paragraph-count checks. After post-generation quality control, \dataset contains \numofdata English prompt variants spanning the five selected
writing-related task categories.

\dataset is not intended to reproduce every idiosyncratic pattern found in non-native writing. Rather, it operationalizes fluency using a theory-grounded taxonomy that decomposes this inherently difficult-to-define construct into five established and complementary dimensions: Content, Organization, Vocabulary, Language Use, and Mechanics. Taken together, these dimensions provide a robust and comprehensive representation of fluency that extends beyond surface-level grammatical correctness to include higher-level rhetorical, lexical, and organizational properties.

\subsection{Evaluating Translated Prompts}
\label{sec:translation-evaluation}

To analyze the effectiveness of our approach for generating prompts with controlled fluency levels, we evaluate the translated prompts along two dimensions: language fluency and semantic preservation. For language fluency, we apply the evaluation pipeline described above in Section ~\ref{Fluency Measurement} to assess the \textbf{Surface Form Features} and \textbf{Rhetorical and Lexical Qualities} of each translated prompt---i.e., does instructing a model to introduce specific aspects of fluency lead to those appearing in the translation? In addition, we consider semantic preservation during translation, as semantic distortion may affect model behavior and downstream task performance. 

We found that higher instructed proficiency levels generally yield corresponding improvements in measured fluency, although cross-dimensional effects remain. Semantic similarity is only weakly associated with instructed fluency, and human evaluation of 100 randomly sampled translations yields a mean semantic-preservation score of 4.13 out of 5, suggesting that the translations generally preserve the original intent. Additionally, instructed levels across all five fluency dimensions show little association with semantic preservation. Detailed methods and results are provided in Appendix~\ref{appendix:evaluation_translated_prompts}.

\section{Quantifying the Impact of Prompt Fluency}

Given two prompts with the same intent, %
but which differ in English fluency,
how different will LLMs' responses be? We evaluate \textit{responses} for their fluency (surface form features and rhetorical and lexical qualities), as well as how well they address the user query (response quality).

\subsection{Experimental Setup}
\label{sec:response_evaluation_setup}
To generate responses to our fluency-controlled prompts, we evaluate 34 LLMs spanning multiple model families and sizes,
including Qwen, Gemma, Llama, Nemotron, OLMo, and others.
The complete list of models is provided in Appendix~\ref{appendix:evaluated_models}. We randomly sample 5,000 English prompt variants from \dataset and use the same sampled set for all models.

Drawing on prior work on LLM-as-judge evaluation
\citep{zheng2023judging,liu2023g}, we develop a response-quality
evaluation protocol using Qwen3.8-27B~\citep{qwen38}. To mitigate the potential inconsistencies caused by varying translation quality in the user prompts, we provide the golden translations obtained earlier in Section \ref{sec:data} to the judging LLM. This design allows the evaluation to focus on differences in model responses rather than artifacts introduced by translation noise. (See Appendix \ref{appendix:response_quality_evaluation} for more details.)

To examine the evaluation protocol, we measure the consistency between the scores of the LLM judge and two human annotators on a random sample of 100 model responses. For overall response quality, human interannotator agreement is Krippendorff's $\alpha = 0.71$,
and the judge's scores achieve a Spearman correlation of
$\rho = 0.55$ with the mean human scores. Given the challenging and subjective nature of evaluating open-ended responses, this moderate positive correlation provides encouraging evidence that the judge captures meaningful aspects of human assessments of response quality.

\subsection{Results}

The effects of prompt fluency are assessed in terms of the generated language and task accuracy. %

\subsubsection{Response Fluency}

LLMs are known to accommodate to users' style \citep{durandard2025language}. For less fluent users, such accommodation is potentially disadvantageous in writing tasks as LLMs may mirror undesired errors. For example, a user asking to help draft an email with a less-fluent prompt may receive a version that has grammatical mistakes.

\begin{figure}[t]
  \includegraphics[width=0.85\columnwidth]{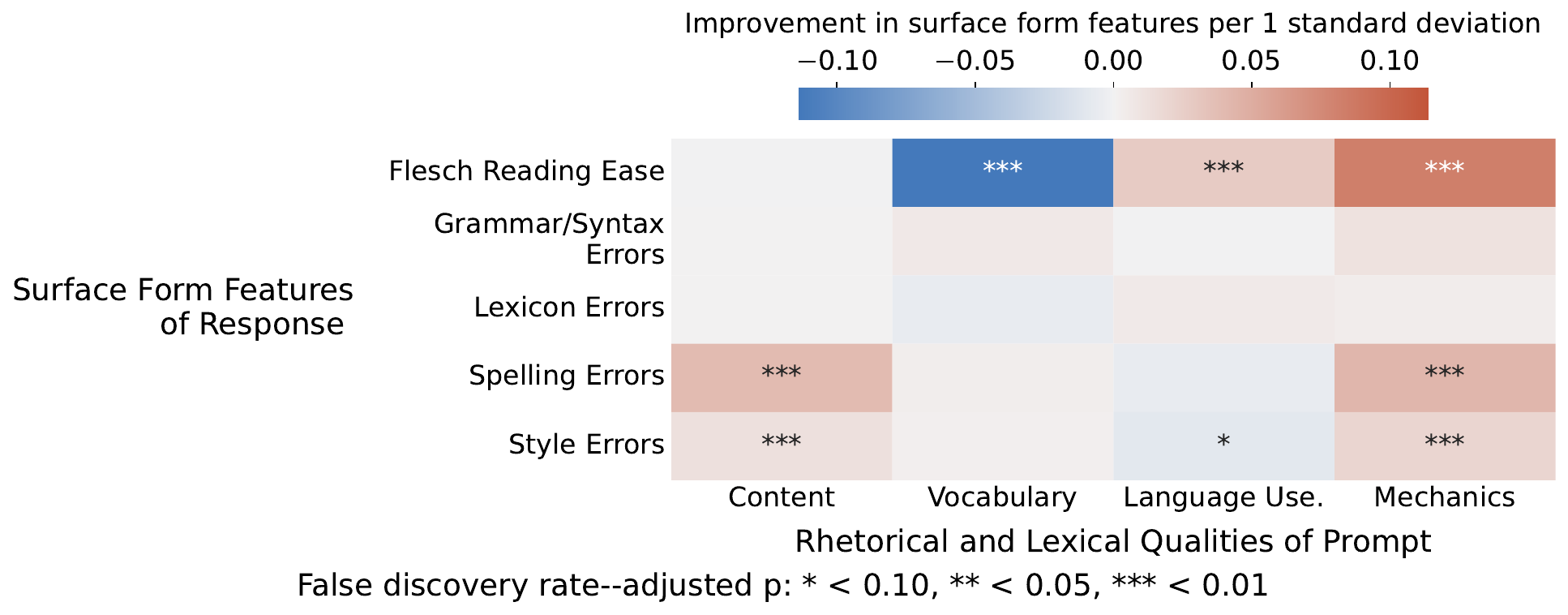}
  \caption{Prompt rhetorical and lexical qualities do not exhibit a systematic impact on model responses' surface form features. This figure presents the regression results between different linguistic properties of responses and different fluency metrics of prompts. %
  }
  \label{fig:responses_linguistic_property}
\end{figure}

\myparagraph{Response Surface Form Features }
To assess the impact of the prompt's fluency, we first fit separate linear regressions to predict the Surface Form Features for each response from the input prompt's rhetorical and lexical quality levels. 

Prompt fluency shows relatively weak associations with the surface form features of LLM responses. As shown in Figure~\ref{fig:responses_linguistic_property}, LLMs did not substantially accommodate to reproduce errors from the prompt. The largest changes were in the reading ease, but changes to other qualities did not produce a systematic trend of introducing more/fewer mechanistic errors. These results suggest that the LLMs we test exhibit relatively stable performance at the lower linguistic level (e.g., few grammar errors) and are largely robust to variations in prompt fluency.
Details of the experiment are provided in Appendix~\ref{appendix:response_surpace_form_feature}.

\begin{figure}[t]
  \includegraphics[width=0.85\columnwidth]{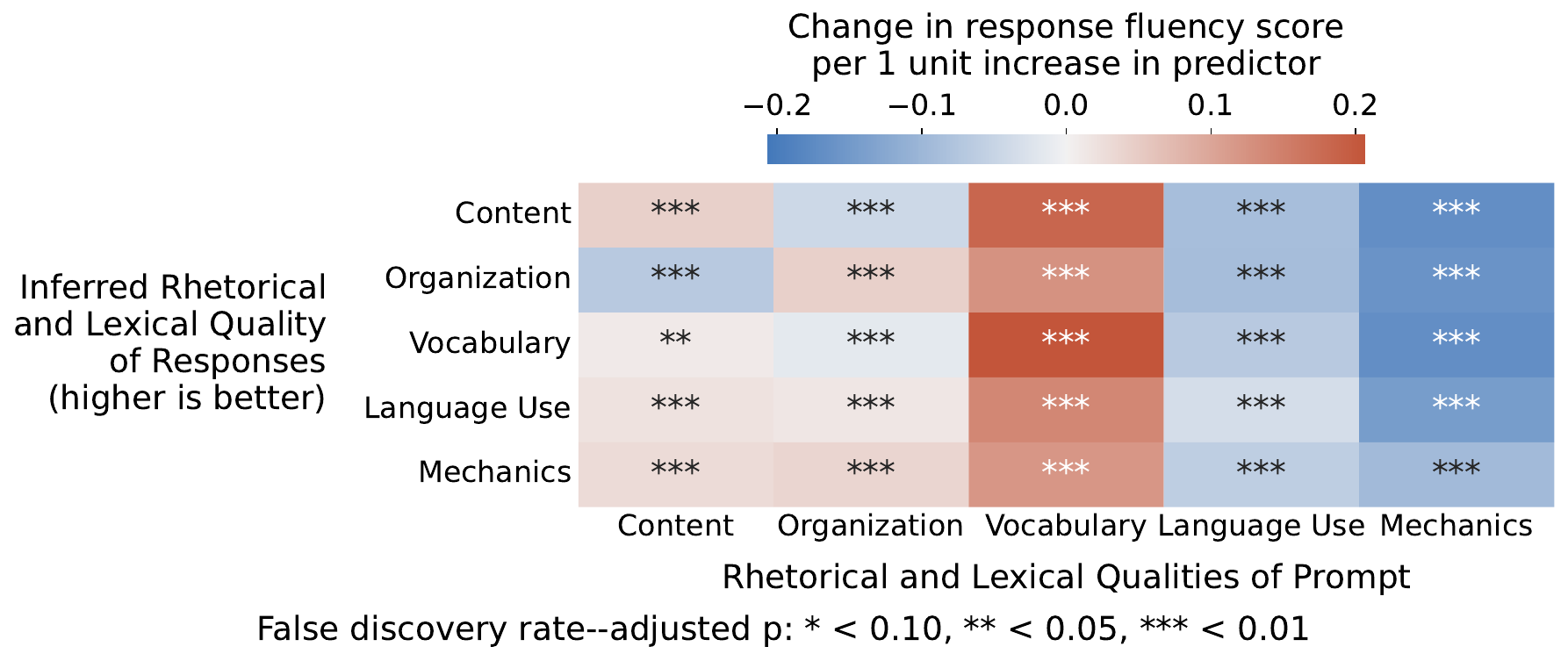}
   \caption{
 The prompt's qualities significantly influence the LLM's response qualities, shown by regression coefficients of the prompt's rhetorical/lexical quality levels (cols) on the response's (row). 
   }
  \label{fig:responses_langugae_fluency_heatmap}
\end{figure}

\myparagraph{Response Rhetorical and Lexical Qualities }
Do the rhetorical and lexical qualities of the prompt influence these qualities in the LLM's response? To answer this, we score responses using the same LLM-as-judge and then, similar to Surface Form Features, we fit a linear regression to predict each of the response's quality levels from the prompt's fluency profile.

LLM responses tend to mirror the rhetorical and lexical qualities of users’ prompts, as shown in Figure~\ref{fig:responses_langugae_fluency_heatmap}.
In particular, higher levels of Content, Organization, and Vocabulary show positive associations with response proficiency, whereas Language Use and Mechanics exhibit negative associations. Thus, users who write with lower levels of Organization or less diverse Vocabulary receive LLM responses matching those qualities. Perceived language proficiency is known to be associated with peer esteem \citep{dev2016relationship} and positive social outcomes \citep[e.g., hiring; ][]{pandey2014better}. Our results suggest that users who are less fluent and attempting to use these models to make up for their gap in fluency will still suffer a penalty due to these LLMs' propagation of errors. Details of the experiment are provided in Appendix~\ref{appendix:response_rhetorical_lexical_quality}.

\subsubsection{Response Correctness}
\label{sec:response_correctness}

\myparagraph{Response Quality }

\begin{figure}[t]
  \includegraphics[width=0.85\columnwidth]{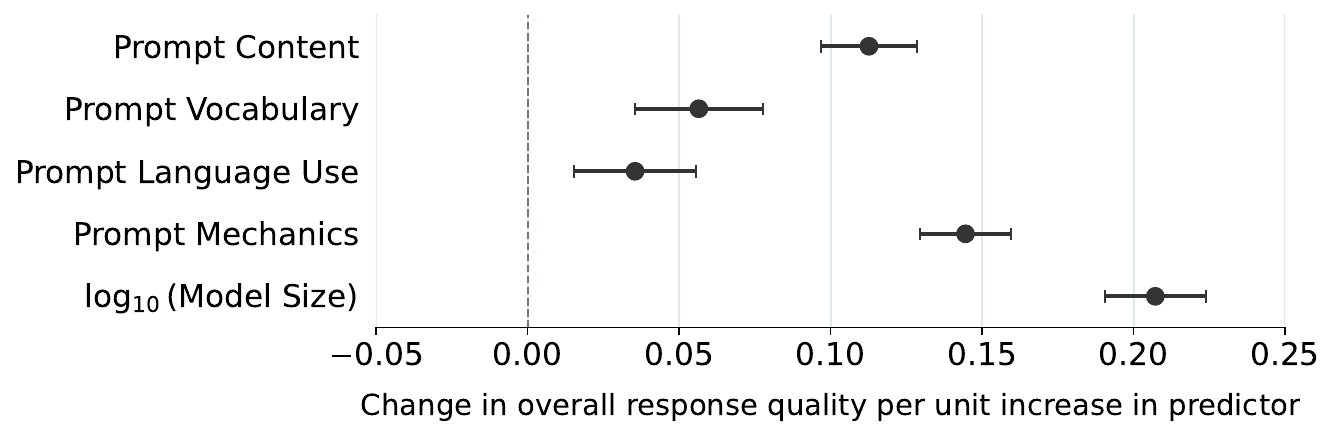}
   \caption{Higher prompt fluency leads to better response
quality, with Content and Mechanics showing larger positive
correlation than Vocabulary and Language Use.
Points represent regression coefficients, and error bars
indicate 95\% confidence intervals.}
  \label{fig:responses_quality_forest}
\end{figure}

Is the correctness of the models' output sensitive to the rhetorical and lexical qualities of the prompt, i.e., do models give less-accurate outputs when the prompt is less fluent? To answer this question, we evaluate using the protocol described in Section~\ref{sec:response_evaluation_setup} and obtain
an overall quality score for each response and analyze
its relationship with prompt fluency using regression.

Prompt fluency has a notable impact on model responses, more fluent prompts generally leading to higher quality outputs, as seen in Figure \ref{fig:responses_quality_forest}. More specifically, a one-point increase in a prompt fluency dimension on the four-point scale is associated with an estimated 0.04--0.14 point increase in overall response quality on a five-point scale. For reference, a tenfold increase in model size (going from a 1B to 10B model) is associated with an approximately 0.21 increase additional response-quality points, which suggests less fluent LLM users experience substantially worse responses, akin to those given by much less capable models. Among all fluency metrics, Content and Mechanics show a more notable impact on response quality compared to Vocabulary or Language Use. One possible explanation is that Content and Mechanics capture how clearly a request is communicated, whereas richer vocabulary and more proficient grammar may provide smaller additional benefits when the intended request is already understandable. Details of the experiment are provided in Appendix \ref{appendix:response_quality_experiment}.

\myparagraph{Semantic Matched Pair Analysis } 

\begin{figure}[t]
  \includegraphics[width=0.8\columnwidth]{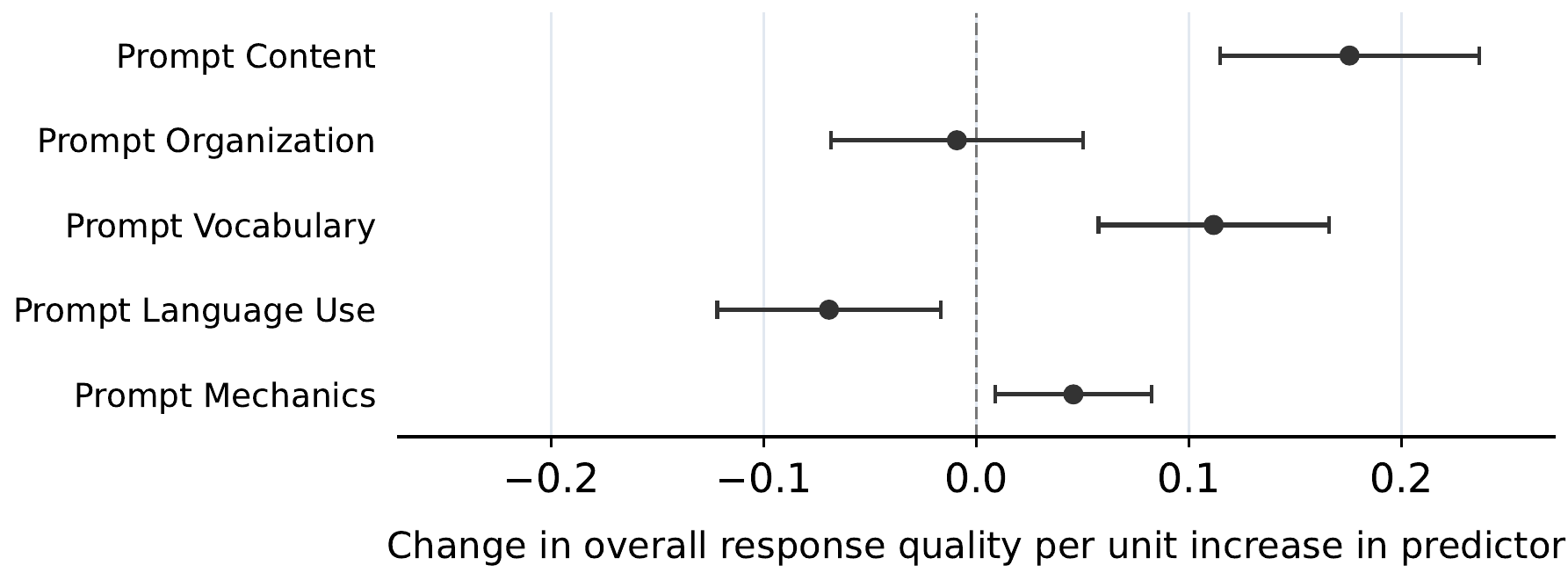}
   \caption{Different prompt fluency dimensions exhibit different
impact on response quality in semantically matched
comparisons. Points represent regression coefficients
relating within-pair differences in each fluency dimension
to differences in overall response quality.}
  \label{fig:mached_pair_zh}
\end{figure}

To further examine the relationship between prompt fluency and response quality while reducing variation in semantic content, we conduct a complementary analysis using semantically matched pairs of Chinese prompts collected from real-world user prompts. For each pair, we translate two prompts with different fluency profiles and then regress differences in overall response quality on differences in the five
prompt fluency dimensions.

Even with similar semantic intent, more fluent prompts generally receive higher-quality responses, as shown in Figure \ref{fig:mached_pair_zh}. Content, Vocabulary, and Mechanics retain positive impact on response quality. Organization has a coefficient close to zero, whereas Language Use shows a negative association. These results further support our previous findings by being broadly consistent with the preceding analysis. For prompts with similar semantic intent, simply improving the fluency level can significantly improve the response quality. This suggests that refining how a request is expressed can help users obtain better responses in everyday interactions with LLMs. Details of the experiment are provided in Appendix~\ref{appendix:matched_pairs}.

\myparagraph{Does prompt fluency impact the output of some tasks more? }

\begin{figure}[t]
  \includegraphics[width=0.85\columnwidth]{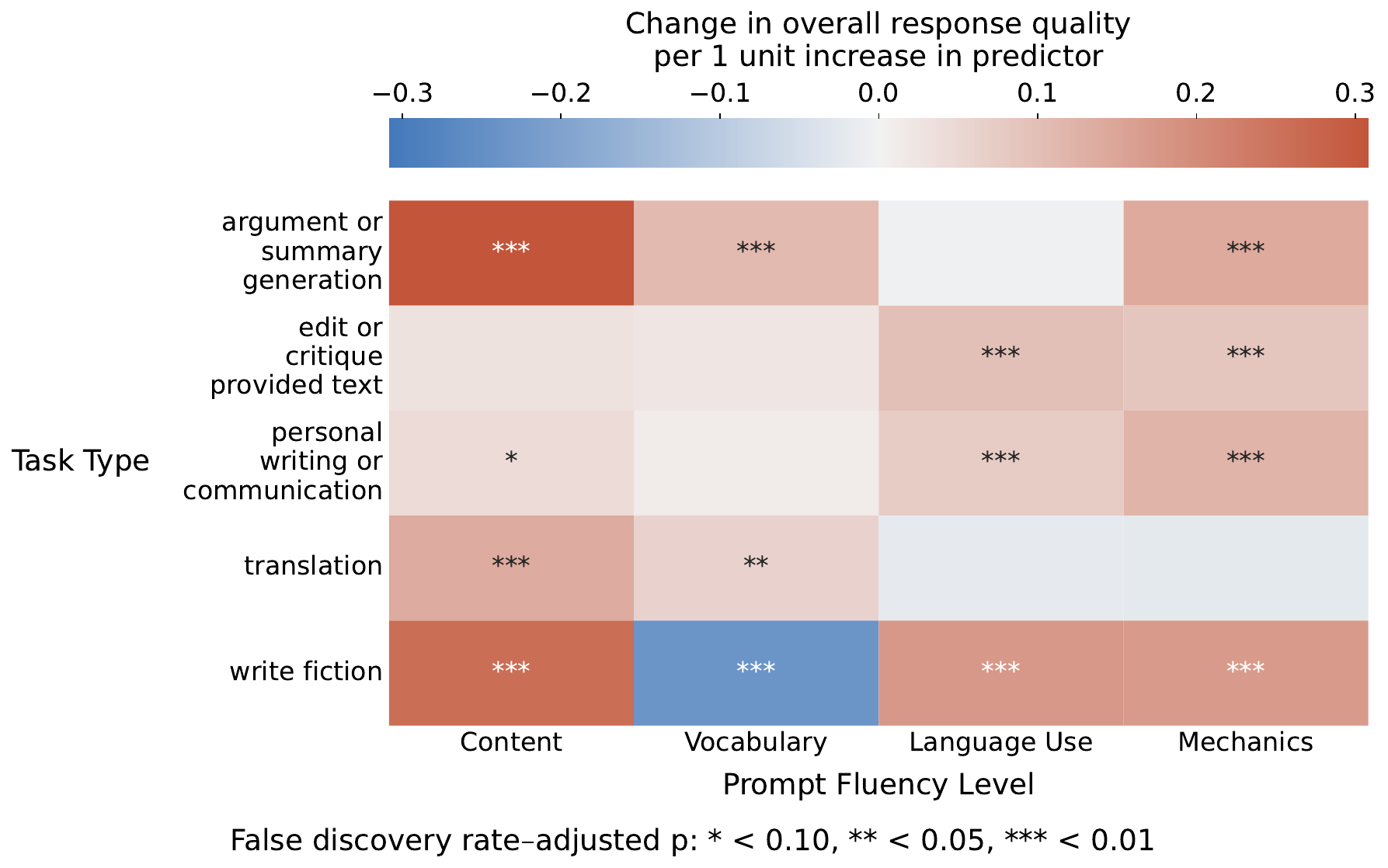}
   \caption{Prompt fluency shows broadly consistent positive associations with response quality across task types, with variation in individual dimensions. Colors indicate coefficients from separate
regressions within each task.}
  \label{fig:responses_quality_heatmap_by_task}
\end{figure}

Given that models produce less fluent and lower quality responses for less fluent prompts, here we assess whether this behavior is more pronounced in certain tasks. We apply a similar regression analysis to measure the relationship between response quality and different prompt fluency dimensions by task type, shown in Figure~\ref{fig:responses_quality_heatmap_by_task}. 

More fluent prompts generally receive higher-quality responses across writing tasks, although the strength of this relationship varies by task types.
Specifically, ``translation" exhibits relatively weak associations with Language Use and Mechanics, while ``editing or critiquing provided text" and ``personal writing or communication" show less sensitivity to Content
and Vocabulary. ``Fiction writing" presents a notable exception to the generally positive pattern, with Vocabulary negatively associated with response quality. Details of the experiment are provided in Appendix~\ref{appendix:task_type_analysis}.

Overall, our results suggest that prompt fluency has a substantial influence on model response quality. More fluent prompts generally lead to responses with higher quality across different task types, although the magnitude of this effect varies slightly depending on the task. Among different fluency dimensions, Content and Mechanics generally exhibit stronger influences on response quality. These findings suggest that the effectiveness of prompts depends not only on their semantic intent, but also on how fluently and coherently users express their intentions. More broadly, our results provide practical implications for real-world interactions with LLMs, indicating that improving prompt fluency may serve as a simple yet effective strategy for obtaining high quality model responses.

\section{Conclusion}

We investigate how the fluency of non-native English prompts shapes the responses that large language models produce. To enable controlled experimentation, we curate \dataset, a collection of \numofdata writing task--related prompts that vary systematically in fluency while preserving semantic intent. In experiments with responses from 34 popular open-source models, we find that LLMs mirror the rhetorical and lexical qualities of the prompt, and their overall response quality is consistently lower for less fluent prompts. More specifically, a one-point increase in a retained fluency dimension on the four-point scale corresponds to an estimated 0.04--0.14 point increase in response quality on the five-point scale. These effects are broadly consistent across writing tasks. Taken together, our findings indicate that non-native English-speaking users of LLMs are likely disadvantaged through their degree of fluency in writing prompts. This disparity in quality is driven less by surface mistakes than by the higher-level discourse and lexical qualities of their prompts, and we hope \dataset enables future work on diagnosing and mitigating this disparity.

\subsection*{AI use statement}

In this work, we used generative AI tools for generating synthetic datasets, assisting with translation, cleaning and reformatting data, designing and providing feedback on research methodology and experiments,  implementing methods, supporting qualitative and thematic data analysis,  creation of artifacts, discovering research topics and identifying gaps, brainstorming, sourcing/searching for information, identifying relevant literature, summarizing and analyzing existing literature, and proposing a title and keywords for the research paper.
We have not used generative AI tools for help developing theoretical models or conceptual frameworks, formulating mathematical claims, providing critical ingredients for proving mathematical claims, assisting in the writing of proofs, proposing or refining hypotheses, interpreting results, formulating questions for surveys or interviews, creating or modifying scientific figures or images, suggesting experimental parameters, formatting references, suggesting a structure for the research paper, or transcribing recordings of research material.
Additionally, we used generative AI tools for creating and editing software code, drafting parts of this research paper, and editing to improve readability. We have reviewed all AI-assisted work, including manually reviewing model outputs and proposed edits to writing. We take responsibility for the final content of this work, including text, claims or artifacts produced with the aid of generative AI.

\subsection*{Reproducibility statement}

To support reproducibility of our results, we will release downloadable source code and the \dataset data upon publication. We also provide a complete description of data processing steps in the main text of the paper and Appendix, including prompts used for data synthesis and response rating.

\bibliography{iclr2027_conference,custom}
\bibliographystyle{iclr2027_conference}

\appendix
\label{sec:appendix}
\section{Appendix}

\subsection{Translated Prompt Synthesizing Details}
\label{appendix:synthesize_prompt}
We use the below prompt to synthesize prompt translations according to specific fluency levels in the ESL Composition Profile.
{\small\begin{lstlisting}[breaklines=true, breakatwhitespace=true]
CONTENT_SPECS = {
4: "Content is rich and fully developed with specific, relevant details that clearly address the topic.",
3: "Content shows adequate knowledge of the topic and generally relevant ideas, though some points lack depth.",
2: "Content shows limited knowledge of the topic, with few supporting details and underdeveloped ideas.",
1: "Content shows very little grasp of the topic; ideas are vague, off-topic, or insufficient to evaluate."}

ORGANIZATION_SPECS = {
4: "Ideas are clearly organized with logical sequencing, clear paragraphing, and strong cohesion.",
3: "Organization is generally clear, though some parts may be loosely connected or uneven in sequencing.",
2: "Organization is weak; ideas are sometimes confused or disconnected and lack clear progression.",
1: "Organization is minimal; ideas are hard to follow and overall structure is unclear or missing."}

VOCAB_SPECS = {
4: "Uses a wide range of vocabulary with accurate and appropriate word choice and idiomatic expressions.",
3: "Uses an adequate range of vocabulary with occasional word-choice errors, but meaning is usually clear.",
2: "Uses a limited range of vocabulary with frequent word-choice problems; meaning is sometimes obscured.",
1: "Vocabulary is very limited, with frequent inappropriate or incorrect word choices; meaning is often unclear."}

LANGUSE_SPECS = {
4: "Uses effective simple and complex sentence constructions with few errors in tense, agreement, and word order.",
3: "Uses mostly correct simple sentences and some complex ones; grammar errors occur but rarely block understanding.",
2: "Shows major problems with grammar and sentence structure; errors are frequent and sometimes obscure meaning.",
1: "Has very little control of sentence construction; grammar errors dominate and communication often breaks down."}

MECH_SPECS = {
4: "Demonstrates mastery of spelling, punctuation, capitalization, and paragraphing with almost no errors.",
3: "Shows occasional errors in spelling, punctuation, capitalization, or paragraphing, but meaning is not obscured.",
2: "Shows frequent errors in spelling, punctuation, capitalization, or paragraphing; meaning may be affected.",
1: "Shows pervasive errors in spelling, punctuation, capitalization, and paragraphing; text is hard to read."}


user_message = f'''
    When producing text, you must simulate the following ESL composition profile:
    Content (level): {content_desc}
    Organization (level): {org_desc}
    Vocabulary (level): {vocab_desc}
    Language use (level): {lang_desc}
    Mechanics (level): {mech_desc}
Translate the following question into English that matches this profile: {original_prompt}
'''
\end{lstlisting}}

\subsection{Translation Examples}

\tref{tab:translation_example} shows examples of the same prompt translated with different uniform rating levels for each quality in instructions (e.g., all rated 4). Prompts retain much of their semantic equivalence, while prompts from instructions with lower ratings have visibly more errors.

\begin{table}[htbp]
\centering
\small
\begin{tabular}{p{0.45cm}p{12.cm}}
\hline
\textbf{Level} & \textbf{Prompt} \\
\hline
1 & Your foreign friend want to introduce Chinese food to classmate, send email ask about information. Please write a reply email, content include: 1. intoduction to Chinese food; 2. recommend a food and explain reason. Word count about 120. \\
2 & Your foreign friend wants to introduce Chinese food to the class and sent an email asking for some information. Please write a reply email, including: 1. an introduction to Chinese food; 2. recommend a dish and explain the reason. Word count around 120. \\
4 & Your foreign friend is going to introduce Chinese cuisine to classmates and has sent an email inquiring about relevant information. Please write a reply email to him, including: 1. An introduction to Chinese cuisine; 2. Recommend a dish and explain the reasons. Word count: around 120 words.\\
\hline
\end{tabular}
\caption{Example translations generated from the same Chinese prompt under different instructed levels. While the semantic intent remains largely preserved, the fluency shows a notable gap across different levels.}
\label{tab:translation_example}
\end{table}

\subsection{Evaluation of Translated Prompts}
\label{appendix:evaluation_translated_prompts}
\begin{figure}[t]
  \centering
  \includegraphics[width=0.8\columnwidth]
  {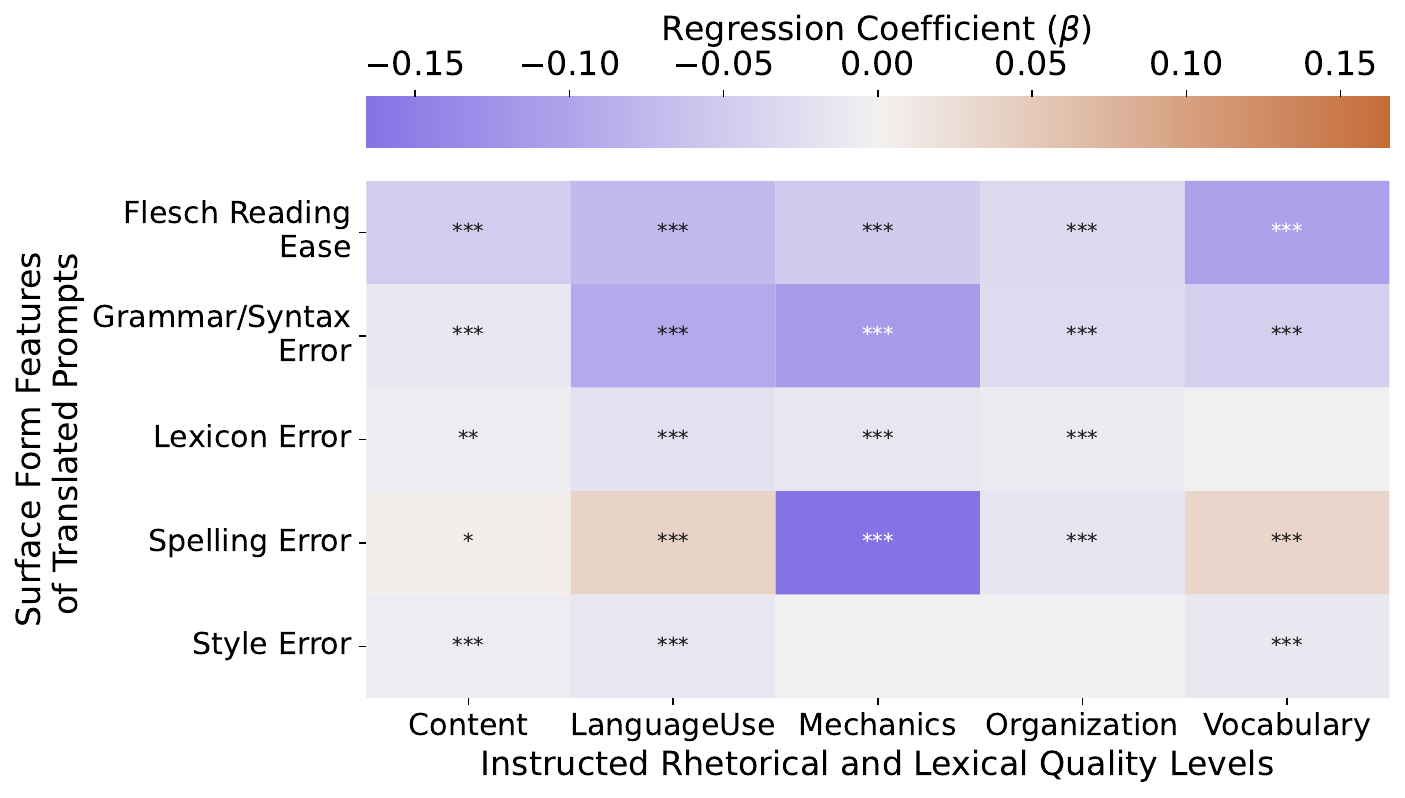}
  \caption{
  Standardized regression coefficients relating instructed
  proficiency levels to surface form features of translated
  prompts. Negative coefficients indicate fewer errors for
  error metrics; positive coefficients indicate greater
  reading ease for readability.
  Stars denote statistical significance:
  * $p<0.1$, ** $p<0.05$, *** $p<0.01$.
  }
  \label{fig:prompt_language_error}
\end{figure}

\myparagraph{Surface Form Features}
Higher instructed proficiency levels generally yield fewer
linguistic errors and improved readability as shown in Figure~\ref{fig:prompt_language_error}.
In particular, higher Mechanics levels are associated with
fewer spelling and grammar--syntax errors, while higher
Language Use levels are associated with improved readability.
These relationships are not uniform: higher Vocabulary
levels are associated with more detected spelling errors.

\begin{figure}[t]
  \centering
  \includegraphics[width=0.85\columnwidth]
  {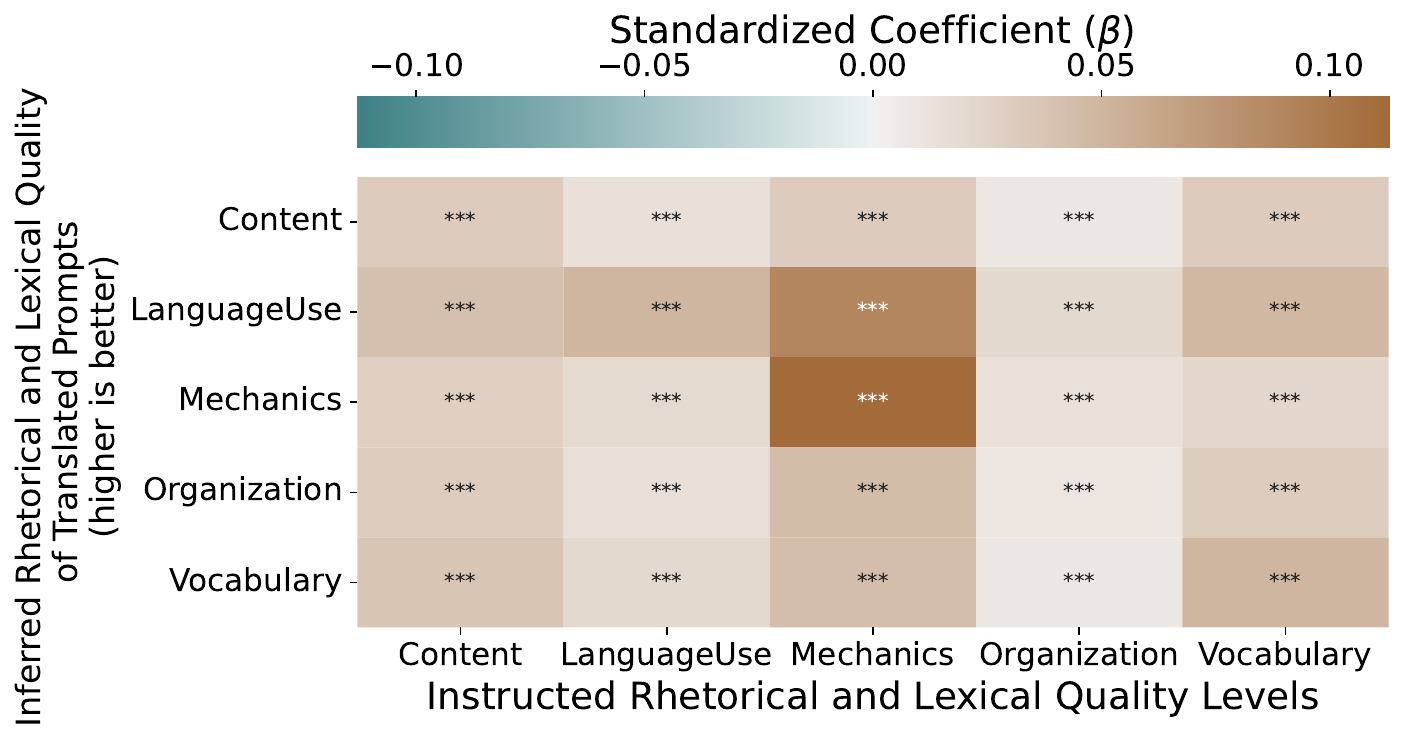}
  \caption{
  Standardized regression coefficients relating the five
  instructed proficiency levels to judged proficiency
  scores of translated prompts. Results show both targeted
  and cross-dimensional associations.
  Stars denote statistical significance:
  * $p<0.1$, ** $p<0.05$, *** $p<0.01$.
  }
  \label{fig:prompt_language_fluency_heatmap}
\end{figure}

\myparagraph{Rhetorical and Lexical Qualities}
Higher instructed levels generally correspond to higher
judged proficiency scores as shown in Figure~\ref{fig:prompt_language_fluency_heatmap}.
The instructed level of a dimension generally has the
strongest association with its corresponding measured
score, particularly for Vocabulary, Language Use, and
Mechanics. Cross-dimensional associations are also
present, indicating that the instructions produce targeted
fluency variation without fully independent control of
the five dimensions.

\myparagraph{Semantic Preservation}
Cross-lingual cosine similarity, measured using
\texttt{BAAI/bge-m3}~\citep{chen2024bge}, varies little
with instructed fluency. The Spearman correlation between
the average instructed level and similarity to the
original Chinese prompt is 0.06.

Human evaluation of 100 randomly sampled translations
yields a mean semantic-preservation score of 4.13 on a
five-point scale, where 1 denotes severe information loss
and 5 denotes no loss. This suggests that the translations
generally preserve the original semantics with limited
information loss on average.
Additionally, Pearson correlations between individual
instructed levels and the same-intent score are presented in
Table~\ref{tab:semantic_preservation_correlations}.
Together, these results support generally preserved
intent and little association between instructed fluency
and semantic preservation, while not ruling out semantic
changes in individual translations.

\begin{table}[htbp]
\centering
\small
\caption{
Pearson correlations between instructed proficiency
levels and the same-intent score.
}
\label{tab:semantic_preservation_correlations}
\begin{tabular}{@{}lr@{}}
\toprule
\textbf{Instructed dimension} & \textbf{Correlation} \\
\midrule
Content      & $-0.112$ \\
Organization & $0.073$  \\
Vocabulary   & $-0.063$ \\
Language Use & $-0.010$ \\
Mechanics    & $-0.015$ \\
\bottomrule
\end{tabular}
\end{table}

\subsection{Evaluated Models}
\label{appendix:evaluated_models}

Table~\ref{tab:evaluated_models} lists the 34 models
evaluated in our experiments.
All models receive the same set of 5,000 English prompt
variants randomly sampled from \dataset, and their
responses are assessed using the same evaluation pipeline.

\begin{table}[t]
\centering
\caption{The 34 evaluated models, grouped by model family.
Version and release identifiers are retained where applicable.
IT denotes instruction-tuned models.}
\label{tab:evaluated_models}
\small
\setlength{\tabcolsep}{8pt}
\renewcommand{\arraystretch}{1.05}
\begin{tabular}{@{}ll@{}}
\toprule
\textbf{Model family} & \textbf{Model name} \\
\midrule
Nemotron
& NVIDIA Nemotron Nano 9B v2 \\
& Llama 3.3 Nemotron Super 49B v1.5 \\
\midrule
OLMo
& OLMo 2 13B Instruct (1124) \\
& Olmo 3 7B Instruct \\
& Olmo 3.1 32B Instruct \\
\midrule
Phi
& Phi-4 \\
\midrule
Qwen
& Qwen3-30B-A3B-Instruct-2507 \\
& Qwen3-32B \\
& Qwen3-4B \\
& Qwen3-8B \\
& Qwen3.5-0.8B \\
& Qwen3.5-35B-A3B \\
& Qwen3.5-4B \\
& Qwen3.5-9B \\
& Qwen3.6-27B \\
& Qwen3.6-35B-A3B \\
& Qwen3.8-27B \\
\midrule
Seed-OSS
& Seed-OSS-36B-Instruct \\
\midrule
SmolLM
& SmolLM3-3B \\
\midrule
Apertus
& Apertus 70B Instruct (2509) \\
& Apertus 8B Instruct (2509) \\
\midrule
Aya
& Aya Expanse 32B \\
\midrule
EXAONE
& EXAONE 4.0 32B \\
\midrule
Gemma
& Gemma 4 12B IT \\
& Gemma 4 26B-A4B IT \\
& Gemma 4 31B IT \\
& Gemma 4 E2B IT \\
& Gemma 4 E4B IT \\
\midrule
Granite
& Granite 4.2 30B \\
& Granite 4.2 3B \\
& Granite 4.2 8B \\
\midrule
Llama
& Llama 3.3 70B Instruct \\
\midrule
Ministral
& Ministral 8B Instruct (2410) \\
\midrule
Mistral
& Mistral Small 3.2 24B Instruct (2506) \\
\bottomrule
\end{tabular}
\end{table}

\subsection{Fluency Evaluation}
\label{appendix:fluency_evaluation}
We use the below prompt to evaluate the fluency of LLM responses using the ESL Composition Profile.
{\small\begin{lstlisting}[breaklines=true, breakatwhitespace=true]
user_message = f'''
Evaluate writing strictly according to the provided rubric:{rubric}.
Evaluate the following text on 5 dimensions: Content, Organization, Vocabulary, LanguageUse, Mechanics.
Each score must be an integer from 1 to 4.
'''
\end{lstlisting}}

\subsection{Response Quality Evaluation}
\label{appendix:response_quality_evaluation}
We use the below prompt to evaluate the quality of LLM responses to instructions of varying fluency.
\begin{lstlisting}[
  basicstyle=\small\ttfamily,
  breaklines=true,
  breakatwhitespace=true,
  columns=fullflexible,
  keepspaces=true,
  showstringspaces=false
]
You are an impartial evaluator of AI assistant responses.
Evaluate the assistant response relative to the user's question.
Score each evaluation dimension independently.
Be objective, consistent, and appropriately critical.
Do not let the score for one dimension automatically determine another dimension.
Output only valid JSON.
For each dimension, provide a brief justification before assigning the score.
Each justification must be concise and no more than 30 words.
\end{lstlisting}

\noindent\textbf{User message}
\begin{lstlisting}[
  basicstyle=\small\ttfamily,
  breaklines=true,
  breakatwhitespace=true,
  columns=fullflexible,
  keepspaces=true,
  showstringspaces=false
]
Evaluate the assistant response on the following five dimensions:

1. Helpfulness: How useful the response is in helping the user achieve their goal.
2. Relevance: How directly the response addresses the user's question without unnecessary or off-topic content.
3. Accuracy: How factually correct and logically sound the response is.
4. Creativity: How effectively the response demonstrates originality, insight, or creative problem-solving when appropriate.
5. LevelOfDetail: Whether the response provides an appropriate amount of specific and useful detail.

Also provide an Overall score reflecting the overall quality of the response.
The Overall score should be a holistic judgment and should not simply be the arithmetic average of the five dimension scores.

For every dimension and the Overall score, use the same rating scale:
1 = Very Poor
2 = Poor
3 = Average
4 = Good
5 = Very Good

Evaluate each dimension independently based only on that dimension.

Output exactly in the following JSON format:
{
  "Helpfulness": {
    "reason": "brief justification",
    "score": 1
  },
  "Relevance": {
    "reason": "brief justification",
    "score": 1
  },
  "Accuracy": {
    "reason": "brief justification",
    "score": 1
  },
  "Creativity": {
    "reason": "brief justification",
    "score": 1
  },
  "LevelOfDetail": {
    "reason": "brief justification",
    "score": 1
  },
  "Overall": {
    "reason": "brief justification",
    "score": 1
  }
}

[Question]
{reference_translation}

[The Start of Assistant's Answer]
{model_response}
[The End of Assistant's Answer]
\end{lstlisting}

\subsection{Additional Experimental Details}

\subsubsection{Response Surface Form Features}
\label{appendix:response_surpace_form_feature}
For each response surface form measure, we estimate
\[
Y = \alpha + \beta_C C + \beta_V V
+ \beta_L L + \beta_M M
+ \boldsymbol{\gamma}^{\top}\mathbf{X} + \epsilon,
\]
where $C$, $V$, $L$, and $M$ denote prompt Content, Vocabulary,
Language Use, and Mechanics scores. $\mathbf{X}$ includes
log-transformed model size, prompt length, response length,
and model-family fixed effects. The outcome and all continuous
predictors are standardized. We exclude Organization because
its VIF exceeded 5, to reduce collinearity. The responses used in this experiment were generated by 34 models mentioned in Appendix~\ref{appendix:evaluated_models}.

\subsubsection{Rhetorical and Lexical Quality of Response }
\label{appendix:response_rhetorical_lexical_quality}
We fit a separate regression for each metric of response
rhetorical and lexical quality:
\[
Y_k = \alpha_k + \beta_{k,C} C + \beta_{k,O} O
+ \beta_{k,V} V + \beta_{k,L} L + \beta_{k,M} M
+ \epsilon_k,
\]
where $Y_k$ is the response score for metric $k$, and
$C$, $O$, $V$, $L$, and $M$ are the prompt's Content,
Organization, Vocabulary, Language Use, and Mechanics scores,
respectively. The responses used in this experiment were generated by Qwen3-14B, Llama-3.1-8B, and GPT-OSS-20B.

\subsubsection{Response Quality}
\label{appendix:response_quality_experiment}
We fit an OLS regression:
\[
Q = \alpha + \beta_C C + \beta_V V
+ \beta_L L + \beta_M M + \beta_S \log_{10}(S)
+ \boldsymbol{\gamma}^{\top}\mathbf{X} + \epsilon,
\]
where $Q$ is overall response quality on a five-point scale;
$C$, $V$, $L$, and $M$ are the prompt's Content, Vocabulary,
Language Use, and Mechanics scores on a four-point scale;
and $S$ is model size in billions of parameters.
$\mathbf{X}$ includes prompt length, response length,
and model-family fixed effects. We exclude Organization to reduce collinearity because we identify its VIF $>5$. The responses used in this experiment were generated by 34 models mentioned in Appendix~\ref{appendix:evaluated_models}.

\subsubsection{Semantically Matched Pair Analysis}
\label{appendix:matched_pairs}
For semantically matched prompt pairs, we fit:
\[
\Delta Q = \alpha + \beta_C \Delta C + \beta_O \Delta O
+ \beta_V \Delta V + \beta_L \Delta L
+ \beta_M \Delta M + \epsilon,
\]
where $\Delta$ denotes the difference within each pair, $Q$ is overall response quality, and $C$, $O$, $V$, $L$, and $M$ are the five prompt fluency scores. We estimate the regression using OLS on unstandardized
differences and report 95\% confidence intervals. The responses used in this experiment were generated by Qwen3-14B, Llama-3.1-8B, and GPT-OSS-20B.

\subsubsection{Analysis by Task Type}
\label{appendix:task_type_analysis}
For each task category $t$, we fit a separate OLS regression:
\[
Q = \alpha_t + \beta_{t,C} C + \beta_{t,V} V
+ \beta_{t,L} L + \beta_{t,M} M
+ \boldsymbol{\gamma}_t^\top \mathbf{X} + \epsilon,
\]
where $Q$ is overall response quality; $C$, $V$, $L$, and $M$
are the four retained prompt fluency scores; and $\mathbf{X}$
contains prompt and response lengths. All variables retain their original scales and we exclude Organization to reduce collinearity because we identify its VIF $>5$. The responses used in this experiment were generated by 34 models mentioned in Appendix~\ref{appendix:evaluated_models}.

\subsection{Computational Environment}
We run all our experiments on 4 NVIDIA-L40S-48GB GPUs. All LLM inferences are powered by vLLM 0.5.4 \cite{kwon2023efficient}, Huggingface Transformers 4.43.3 \cite{wolf-etal-2020-transformers} and PyTorch 2.4.0 \cite{NEURIPS2019_9015} on a CUDA 12.4 environment. Temperatures are set to 0.0 to minimize the effect of randomness.

\subsection{Licenses}
All data and code will be publicly released under the CC BY-SA 4.0 license.

\end{document}